\documentclass[10pt,twocolumn,letterpaper]{article}

\usepackage[pagenumbers]{cvpr} 

\usepackage{graphicx}
\usepackage{amsmath}
\usepackage{amssymb}
\usepackage{booktabs}

\usepackage{times}  
\usepackage{helvet}  
\usepackage{courier}  
\usepackage[hyphens]{url}  
\usepackage{caption} 
\usepackage{pifont}
\DeclareCaptionStyle{ruled}{labelfont=normalfont,labelsep=colon,strut=off} 
\usepackage{multirow}
\usepackage{setspace}
\usepackage{bbding}
\usepackage{epsfig}
\usepackage{float}
\usepackage{color}
\usepackage{ulem}

\newcommand{\hupar}[1]{\vspace{12pt} \noindent \textbf{#1}\quad}
\newcommand{\tabincell}[2]{\begin{tabular}{@{}#1@{}}#2\end{tabular}}

\usepackage[pagebackref,breaklinks,colorlinks]{hyperref}

\usepackage[capitalize]{cleveref}
\crefname{section}{Sec.}{Secs.}
\Crefname{section}{Section}{Sections}
\Crefname{table}{Table}{Tables}
\crefname{table}{Tab.}{Tabs.}

\def\cvprPaperID{5331} 
\def\confName{CVPR}
\def\confYear{2022}

\begin{document}

\title{Spike Stream Denoising via Spike Camera Simulation}

\author{Liwen Hu\textsuperscript{\rm 1,2}, 
Lei Ma\textsuperscript{\rm 1,2}\thanks{Corresponding author.},
Zhaofei Yu\textsuperscript{\rm 1,2},
Boxin Shi\textsuperscript{\rm 1,2,3},
Tiejun Huang\textsuperscript{\rm 1,2,3}\\
\textsuperscript{\rm 1}National Engineering Research Center of Visual Technology (NERCVT), Peking University\\
\textsuperscript{\rm 2}Beijing Academy of Artificial Intelligence\\
\textsuperscript{\rm 3}Institute for Artificial Intelligence, Peking University\\
}

\twocolumn[{
\renewcommand\twocolumn[1][]{#1}
\maketitle
\begin{center}
    \captionsetup{type=figure}
    \includegraphics[width=0.95\textwidth]{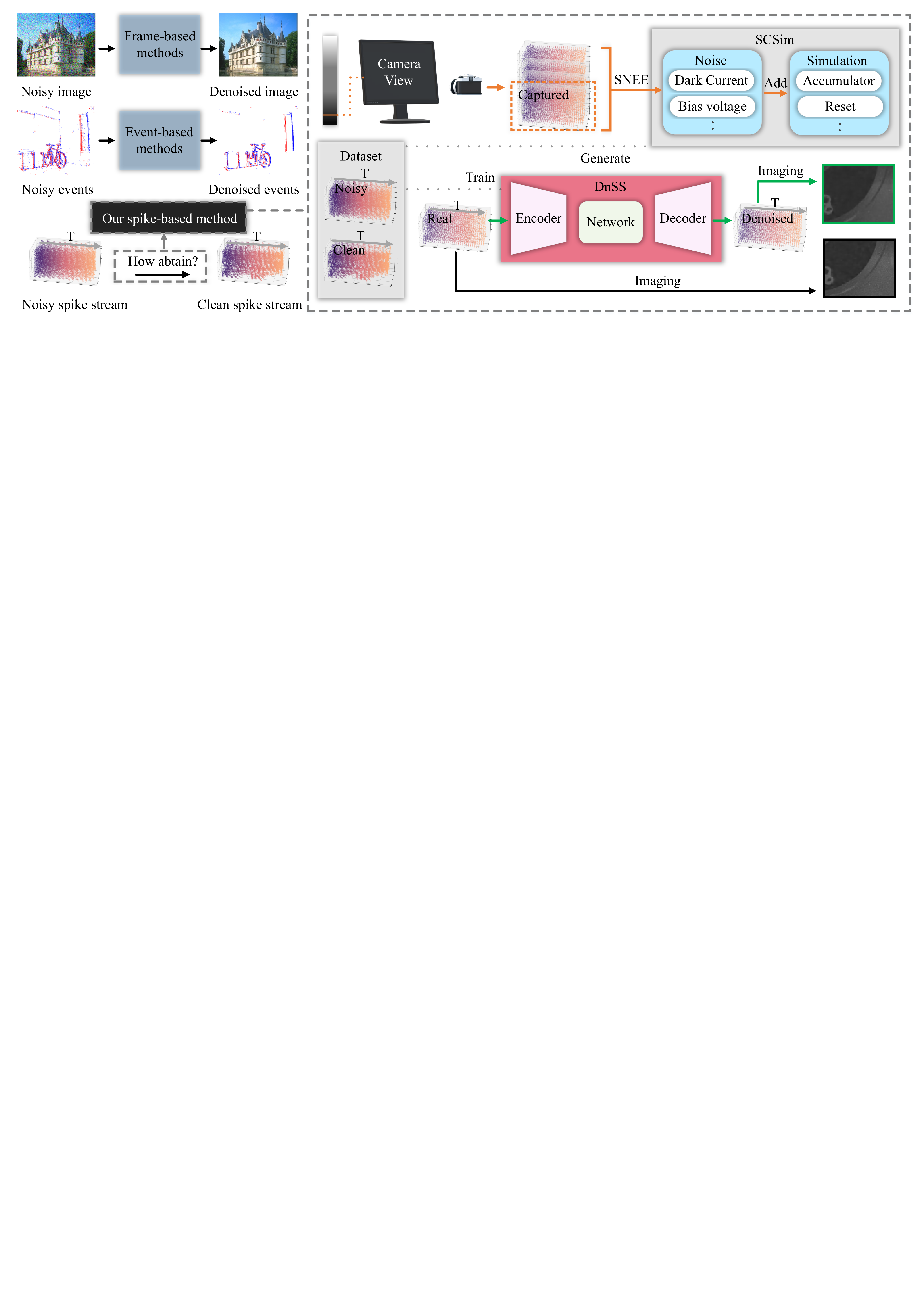}
    \captionof{figure}{Overview of real spike stream denoising which records a high-speed disk with 2000 revolutions per
minute. Based on spike camera simulation (SCSim) where noise variables are estimated by using the spike-based noise evaluation equation (SNEE), we systematically explore spike stream denoising and a tailored framework (DnSS) is proposed.
}\label{teasor}
\end{center}
}]
\maketitle
\begin{abstract}
As a neuromorphic sensor with high temporal resolution, the spike camera shows enormous potential in high-speed visual tasks. However, the high-speed sampling of light propagation processes by existing cameras brings unavoidable noise phenomena.
Eliminating the unique noise in spike stream is always a key point for spike-based methods. No previous work has addressed the detailed noise mechanism of the spike camera. 
To this end, we propose a systematic noise model for spike camera based on its unique circuit. In addition, we carefully constructed the noise evaluation equation and experimental scenarios to measure noise variables. Based on our noise model, the first benchmark for spike stream denoising is proposed which includes clear (noisy) spike stream. Further, we design a tailored spike stream denoising framework (DnSS) where denoised spike stream is obtained by decoding inferred inter-spike intervals. Experiments show that DnSS has promising performance on the proposed benchmark. Eventually, DnSS can be generalized well on real spike stream.
\end{abstract}

\section{Introduction}
High-speed scenes have always been a major challenge for some vision applications such as self-driving cars \cite{2006real-time}. Images captured by conventional cameras with low frame rate suffer from motion blur due to long-time integration of light, which greatly limits the performance of frame-based approaches. The emergence of spike camera \cite{spikecamera,spikecode} provides a new perspective for applications in high-speed scenes. The  bio-inspired spike camera not only has the advantage of high temporal resolution (40000 Hz), but also can report per-pixel luminance intensity by firing spikes. Recently, it has shown enormous potential for high-speed visual tasks such as reconstruction \cite{rec0,rec1,rec2,rec3,rec4, rec5}, optical flow estimation \cite{2022scflow}, and depth estimation \cite{2022spkTransformer}. {Different from event cameras \cite{dvs1, dvs2, dvs3, dvs4, dvs5} and traditional cameras}, the spike camera circuit can introduce unique noise into spike stream during the sampling process. Eliminating noise in spike stream is a key issue for spike-based methods \cite{rec0, rec3, rec5}.
\\\indent 
At present, research on spike stream denoising is still in its infancy. 
{One of the main challenges is the frame-based and event-based methods cannot be directly applied to the spike stream with unique data modality \cite{spikecamera, 2022scflow, rec3}.} Although we can reconstruct clear image sequences first, and then synthesize the spike stream with a noise-free spike camera model, such reconstruction-based denoising pipeline depends on the performance of the reconstruction methods. Traditional reconstruction methods can introduce
extra noise and blur while high-quality reconstruction methods would cost extra processing efforts. Instead of adopting a reconstruction-based pipeline, we should fully utilize features of spike stream to design a tailored denoising method. 
Another challenge is the lack of datasets to evaluate the performance of denoising methods. We can use spike camera simulator \cite{2022scflow, rec4, rec3, 2021spk_sim} to generate noisy spike stream and clean spike stream, but noise models in existing works are unrealistic. In order to benchmark denoising methods for spike camera accurately, we need to delve deeper into its distinct noise
mechanism.
In this paper, as shown in Fig.~\ref{teasor}, we propose a realistic simulation method for spike cameras, SCSim, where its circuits-level noise is modeled. Further, we set up the noise measurement experiment and evaluate statistics of the noise variable by building the relationship between spike stream and luminance intensity. Based on our noise model, the first synthetic  benchmark is introduced to evaluate denoising methods for spike camera. To remove the noise in spike stream, we design a tailored spike stream denoising framework, DnSS, where inter-spike interval (ISI \cite{spikecamera}) in clean spike stream as the supervised signal and the denoising spike stream can be obtained by decoding inferred ISI sequences.
We compare DnSS and the reconstruction-based denoising pipeline. The experiment shows that lightweight DnSS achieves the best performance. Our main contributions are summarized as follows:
\begin{itemize}
\item[$\bullet$] \textbf{A spike camera simulation method:} We analyze the basic principles of spike camera circuit implementation in detail and model its distinct noise.
\item[$\bullet$] \textbf{A measurement method for spike camera noise:} We propose the Spike-based Noise Evaluation Equation (SNEE) to establish the relationship between noise and spike stream, and we collect real spike stream and use SNEE to evaluate the statistics of noise.

\item[$\bullet$] \textbf{Datasets for spike stream denoising:} Based on SCSim and evaluated statistics of noise variables, we propose two datasets for spike stream denoising.
\item[$\bullet$] \textbf{A spike stream denoising framework:} In DnSS, the ISI in clean spike stream can be inferred based on refine module (RF), and then we obtain 
the denoised spike stream by using multi-stage update strategy (MUS) to decode the inferred ISI.
\end{itemize}

\section{Related Work}
\subsection{Frame-based and Event-based Denoising}
\hupar{Image Denoising}
In the early years, hand-designed filters were used for image denoising e.g., non-local means (NLM) algorithm \cite{2005NLM}, and the block-matching based 3D filtering (BM3D) algorithm \cite{2007BM3D} which use a stack of non-local similar patches. However , the difficulty of setting the hyper-parameters of these methods will significantly affect the performance. With the development of deep learning, dncnn\cite{2017dncnn} proposes first denoising neural network which focus on learning a latent mapping from the noisy image to the clean image. After that, numerous denoising networks are proposed to further improve the performance \cite{2018dn_cnn,2018dn_Ffdnet,2019dn_blind,2019dn_Self_guided}.  it is challenging and expensive to collect plenty of training pairs for supervised denoising. To this end, some self-supervised denoising models \cite{2018Noise2noise,2019Noise2self, 2019Noise2void, 2020Noisier2noise, 2020udn_unpair, 2019udn_hq, Neighbor2Neighbor} are proposed.
\cite{2018Noise2noise} introduced Noise2Noise to train a deep denoiser with multiple noisy observations of the same scenes. \cite{2019Noise2void} and \cite{2019Noise2self}  use only one noisy observation per scene to train deep denoisers. By generating sub-sampled paired images with random neighbor sub-samplers as training image pairs, \cite{Neighbor2Neighbor} can denoise images more effienctly.

\hupar{Event Denoising}
Since event stream have the unique data modality, event denoising is treated as a classification problem i.e., distinguishing noise event and real activity event. \cite{2015_ev_dn_filter} proposed a filter to tackle the problem of increased memory requirements for events denoising. \cite{2021_ev_dn_filter} used another storage technique for events and their timestamps to utilize less memory space. \cite{2020_ev_dn_density} proposed a density matrix to denoising event stream in which each arriving event is projected into its spatio-temporal region. \cite{2020_ev_dn_epm} introduced a reliable dataset, DVSNOISE20 by deriving an event probability mask using APS frames and IMU motion data. Based on DVSNOISE20, \cite{2020_ev_dn_epm} and \cite{2021_ev_dn_zoom} proposed a CNN and U-net network to filter DVS noises, respectively.
\subsection{Spike-based Applications and Simulation}
The spike camera shows potential in high-speed visual tasks. Based on the statistical characteristics of spike stream, \cite{spikecamera} first reconstruct high-speed scenes. \cite{rec1, rec2} and \cite{rec3, rec4} respectively use SNN and convolutional neural network to reconstruct high-speed images from a spike stream, which greatly improved the reconstruction quality.
\cite{sup0} present super-resolution framework for the spike camera and recover external scenes with both high temporal and high spatial resolution from spike stream. \cite{2022scflow} present a deep learning pipeline to estimate optical flow in high-speed scenes from spike stream.  \cite{2022spkTransformer} can clearly estimate the depth of high-speed scenes by introducing spike-based transformer. High-speed scenes datasets for spike camera are mainly generated through simulation. \cite{rec3} first convert interpolated image sequences with high frame rate into spike stream. \cite{rec4,2021spk_sim} add simple random noise to the ideal spike camera model. \cite{2022scflow} present the spike camera simulator (SPCS) combining simulation function and rendering engine tightly.

\section{Spike Camera Noise Simulation}
Based on the circuit principle of spike camera, we model and measure the joint interference of various noises for the generation of spike stream. We start by introducing a noise-free spike camera model in Section 3.1. And then we propose a practical model along with the noises in Section 3.2. To measure statistics of noise, a reasonable experiment is designed. Finally, we evaluate the spike camera simulation method (SCSim) in Section 3.3.
\subsection{Ideal Spike Camera Model}
\begin{figure}[htbp]
\includegraphics[width=\linewidth]{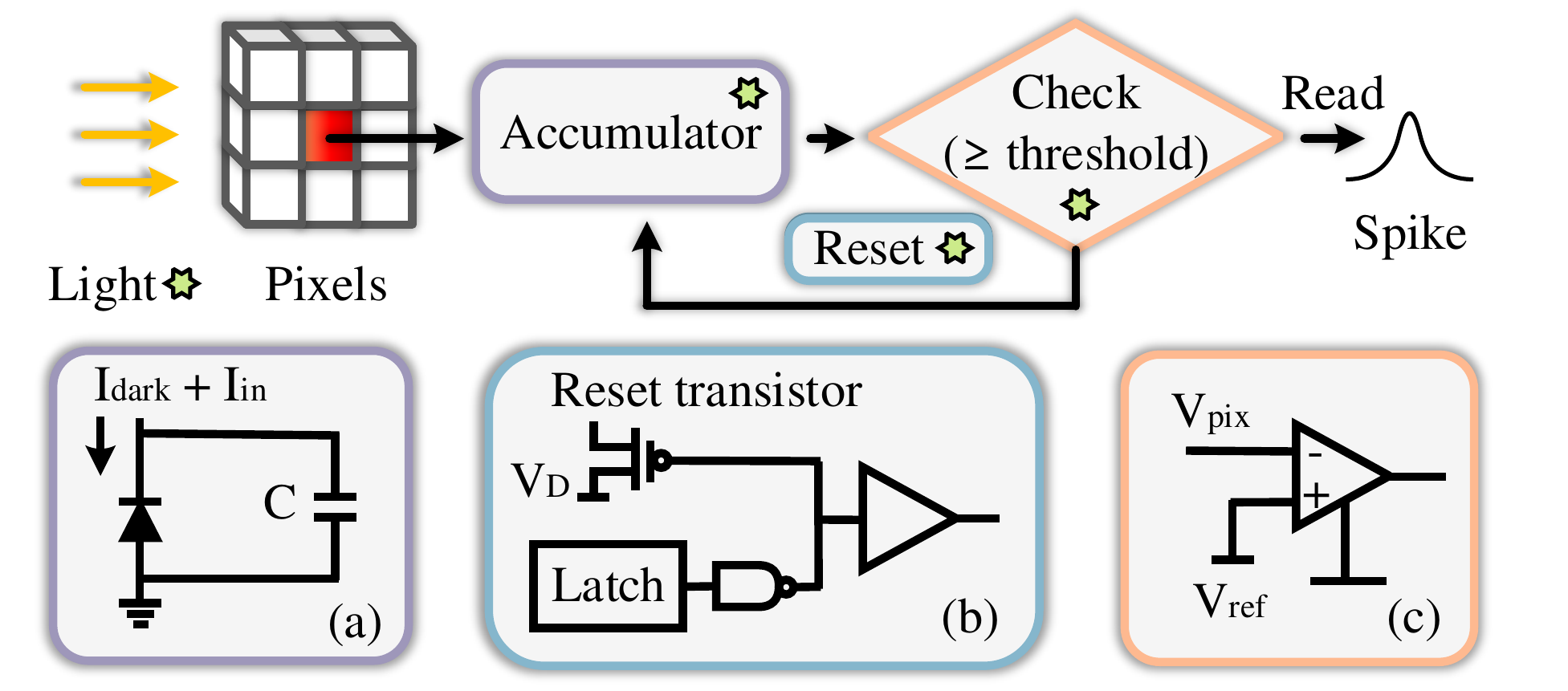}
\centering
\caption{The principle and circuit of spike camera. (a) is the work principle of spike camera. (b) is the pixel circuit of spike camera. (c) is a specific operation process. ``Star" means that we have considered the corresponding noise.}\label{spike_camera_model}
\end{figure}

The spike camera mimicking the retina fovea consists of an array of $H \times W$ pixels and can report per-pixel luminance intensity  by firing asynchronous spikes. Specifically, as shown in Fig.~\ref{spike_camera_model}, each pixel on the spike camera sensor accumulates incoming light independently and persistently. At time $t$, for pixel  $(x, y)$, if the accumulated brightness arrives a fixed threshold  $\phi$ (as (1)), a spike is fired and then the accumulated brightness can be reset to 0.
   \begin{eqnarray}
    {A}(x, y, t) = \int_{t_{x, y}^{\rm pre}}^{t} {I_{in}}(x, y, \tau) d\tau\ \geq \phi, 
    \end{eqnarray}
    where  $x, y\in \mathbb{Z}, x \leq H, y \leq W$, $  {A}(x, y, t)$  is the accumulated brightness at time  $t$,  ${I_{in}}(x, y, \tau)$ is the input and $t_{x, y}^{\rm pre}$ expresses the last time when a spike is fired at pixel  $(x, y)$  before time  $t$. If  $t$   is the first time to send a spike, then  $t_{x, y}^{\rm pre}$  is set as 0. In fact, due to the limitations of circuit technology, the spike reading times are quantified. Hence, asynchronous spikes are read out synchronously. Specifically, all pixels periodically judge the spike flag at time $n\delta t, n \in \mathbb{Z}$, where $\delta t$  is a short interval of microseconds.     
    Therefore, the output of all pixels forms a  $H \times W$   binary spike frame. As time goes on, the camera would produce a sequence of spike frames, i.e., a  $H \times W \times N$  
binary spike stream and can be mathematically defined as, 
   \begin{eqnarray}
    \begin{aligned}
    &  {S}(x, y, n\delta t) =
    \\&
    \begin{cases}
    1 &\mbox{ if  $\exists t \in \left((n - 1)\delta t, n\delta t\right]$, s.t.  $  {A}(x, y, t) \geq \phi$   }, \\
    0 &\mbox{ if  $\forall t \in \left((n - 1)\delta t, n\delta t\right]$, $  {A}(x, y, t) < \phi$  } \\
    \end{cases} 
    \end{aligned}
    \end{eqnarray}
\subsection{Spike Camera Model with Realistic Noise}
We will systematically introduce the temporal noise and spatial noise in spike camera according to its unique circuit.

\hupar{Temporal Noise}
Temporal noise is a random variation in the signal that fluctuates over time. The temporal noise in spike camera mainly includes shot noise and thermal noise. The shot noise originates from randomness
caused by photon reception and, for pixel $(x, y)$, the probability that $n$ photons are received between time $t$ and $t + \delta t$ is given by the Poisson probability distribution, i.e., 
\begin{equation}
P({ph}(x, y, t) = n) = \frac{\mu_{  {ph}(x, y, t)}}{n!e^{\mu_{  {ph}(x, y, t)}}}
\end{equation}
where $n \in \mathbb{N}$, $  {ph}(x, y, t)$ is a random variable representing the number of received photons from $t$ to $t + \delta t$, and $\mu_{  {ph}(x, y, t)}$ is the expectation of $  {ph}(x, y, t)$.
The random number of photons can affect the luminance intensity. Since $\delta t$ is enough small, we consider the luminance intensity at time $t$ is proportional to the number of photons between time $t$ and $t + \delta t$, i.e., $  {L}(x, y, t) \propto   {ph}(x, y, t)$.
Different from the previous simulator \cite{2022scflow} which uses the ideal luminance intensity $\mu_{{L}(x, y, t)}$ as the input, we sample $  {L}(x, y, t)$ at time $t$ during simulation, i.e.,
\begin{equation}
  {L}(x, y, t) = \mu_{  {L}(x, y, t)}\dfrac{  {ph}(x, y, t)}{\mu_{  {ph}(x, y, t)}}.
\end{equation}
The threshold $\phi$ (as (1)) in the ideal spike camera model also fluctuates over time. We start by writing the ideal threshold in the form of the circuit, i.e,
\begin{equation}
\phi = CV_{d} = C(V_{D} - V_{ref})
\end{equation} The voltage fluctuation in the camera can be caused by the reset transistor, which is affected by temperature
where $V_{D}$ is reset voltage and $V_{ref}$ is reference voltage. As shown in Fig.~\ref{spike_camera_model}(b), the reset transistor affected by temperature can cause voltage fluctuation. We use the random variable ${V}^{T_0}(x,y,t)$ to describe fluctuating voltage in pixel $(x, y)$ at time $t$ and ${V}^{T_0}(x,y,t)$ can be considered to obey Gaussian distribution, i.e., 
\begin{equation}
  {V}^{T_0}(x,y,t) \sim N(0, (\sigma^{T_0})^2), \quad  \sigma^{T_0} = \sqrt{\frac{kT_0}{C}},
\end{equation}
where $\sigma^{T_0}$ is the standard deviation of $  {V}^{T_0}(x, y, t)$, $k$ is Boltzmann constant, $T_0$ is absolute temperature.

\hupar{Spatial Noise}
Spatial noise i.e., fixed-pattern noise, is a random variation in the signal that has nothing to do with time. The spatial noise in spike camera mainly includes conversion rate, dark current, capacitor mismatch and bias voltage. As shown in Fig.~\ref{spike_camera_model}(a), we first develop the input $I_{in}$ based on circuit principles, i.e., ${I_{in}}(x, y, \tau) = \alpha{L}(x, y, \tau)$ where $\alpha$ is the photoelectric conversion rate, $  {L}(x, y, \tau)$   refers to the brightness of pixel  $(x, y)$ at time  $\tau$.
The difference of photodiodes can cause the conversion rate to fluctuate. We assume the conversion rate in pixel $(x, y)$ obeys Gaussian distribution as,
\begin{equation}
  {\alpha}(x, y) \sim N(\mu_{\alpha}, (\sigma_{\alpha}^{S})^2),
\end{equation}
where $\mu_{\alpha}$ and $\sigma_{\alpha}$ contribute to the expectation and variance of $\alpha(x, y)$, respectively.
The dark current is also from photodiodes. The dark current $I_{\rm {dark}}(x, y)$ in pixel $(x, y)$ obeys Gaussian distribution as,
\begin{equation}
I_{\rm {dark}}(x, y) \sim N(\mu_{\rm {dark}}, (\sigma_{\rm {dark}}^{S})^2),
\end{equation}
where  $\mu_{\rm {dark}}$ and $\sigma_{\rm {dark}}^{S}$ contribute to the expectation and variance of $I_{\rm {dark}}(x, y)$. Besides, capacitance difference in each pixel obeys Gaussian distribution and can be written as, 
\begin{equation}
C^{S}(x,y) \sim N(0, (\sigma_{C}^{S})^2),
\end{equation}
where $C^{S}(x,y)$ is the random variable describing the capacitance nonuniformity at pixel $(x, y)$, $\sigma_{C}^{S}$ is the standard deviation of $C^{S}(x,y)$.
The bias voltage $V^{S}(x,y)$ mainly comes from reset voltage nonuniformity in check module as shown in Fig.~\ref{spike_camera_model}(c). We can assume the bias voltage $V^{S}(x,y)$ at pixel $(x, y)$ obeys Gaussian distribution, i.e.,
\begin{equation}
V^{S}(x,y) \sim N(0, (\sigma_{V}^{S})^2),
\end{equation}
where  $\sigma_{V}^{S}$ is the standard deviation of $V^{S}(x,y)$.

\hupar{The Sampling Process with Noise}
We can add the above noise model into the generation of spike stream. At time $t$, for pixel $(x, y)$, the accumulated brightness $  {A}(x, y, t)$ in (1) can be rewritten as,
\begin{equation}
\int_{t_{x, y}^{\rm pre}}^t \!\!\!\!\!\! \alpha(x, y) L(x, y, \tau) + I_{\rm {dark}}(x, y) d\tau \geq \phi(x, y, t),
\end{equation}
where $\phi(x, y, t)$ is the threshold affected by noise.
Further, $\phi(x, y, t)$ can be expressed as,
\begin{equation}
\begin{aligned}
(C + C^{S}(x, y))(V_{d} + V^{T_0}(x, y, t) + V^{S}(x, y)).
\end{aligned}
\end{equation}
The spike stream $  {S}(x, y, n\delta t)$ in (2) can be rewritten as,
   \begin{equation}
    \begin{aligned}
    \begin{cases}
    1 \!\!\!\!&\mbox{ if  $\exists t \in \left((n - 1)\delta t, n\delta t\right]$, s.t.  $  {A}(x,y, t) \geq \phi(x,y,t)$}, \\
    0 \!\!\!\!&\mbox{ if  $\forall t \in \left((n - 1)\delta t, n\delta t\right]$, $  {A}(x,y, t) < \phi(x,y,t)$}.  \\
    \end{cases}
    \end{aligned}
    \end{equation}
\subsection{Noise Measurement}
For simulating noise in spike camera accurately, we need to measure statistics of noise, i.e., expectation and variance. Since statistics of temporal noise is known, we only need to measure spatial noise, i.e., $\alpha(x, y)$, $I_{\rm {dark}}(x, y)$, $V^{S}(x, y)$ and $C^S(x, y)$. We first establish the equation spike-based noise evaluation equation (SNEE) according to the meaning of spike. Further, we built the corresponding experimental scenes to capture real spike stream. Finally, noise variables are estimated by solving the equations of real data.

\hupar{Spike-based Noise Evaluation Equation} A spike fired by pixel $(x, y)$ in time $t$ means that the accumulation at the pixel arrives threshold $\phi(x, y, t)$. Hence, if accumulation would not be reset, the total accumulation at the pixel $(x, y)$ in time $n\delta t$ can be estimated as, 
\begin{equation}
\sum_{S(x, y, i\delta t) = 1}^{i}\!\!\!\!\!\!\!\! \phi(x, y, t_i),
\end{equation}
where $i \in \mathbb{Z} \cap [1, n]  \; s. t.  \; S(x, y, i\delta t) = 1$, $t_i$ express the time when the  i-th spike is fired and  $t_i \in ((i - 1)\delta t, i\delta t]$. According to (11), the total accumulation at pixel $(x, y)$ in time $n\delta t$  also can be expressed as,
\begin{equation}
\int_{0}^{n\delta t}\!\!\! \alpha(x, y) L(x, y, \tau) + I_{\rm {dark}}(x, y) d\tau.
\end{equation}
Further, we have the spike-based noise evaluation equation (SNEE), i.e.,
\begin{equation}
\begin{aligned}
&\sum_{S(x, y, i\delta t) = 1}^i\!\!\!\!\!\!\!\! \phi(x, y, t_i)\\
&=\!\!\!\!\!\!\! \sum_{S(x, y, i\delta t) = 1}^i\!\!\!\!\!\!\!\! (C + C^{S}(x, y))(V_{d} + V^{T_0}(x, y, t_i) + V^{S}(x, y)) \\
&=\int_{0}^{n\delta t}\!\!\! \alpha(x, y) L(x, y, \tau) + I_{\rm {dark}}(x, y) d\tau
\end{aligned}
\end{equation}
We can set experimental scenes to control $L(x, y, \tau)$ and eliminate temporal noise by extending sampling time $n\delta t$. When experimental scene is static and sampling time $n\delta t$ is long enough, (16) can be simplified as,
\begin{equation}
\begin{aligned}
&\sum_{S_k(x, y, i\delta t) = 1}^i\!\!\!\!\!\!\!\! (C + C^{S}(x, y))(V_{d} + V^{S}(x, y)) \\
&=\int_{0}^{n\delta t}\!\!\! \alpha(x, y) \mu_{k} + I_{\rm {dark}}(x, y) d\tau
\end{aligned}
\end{equation}
where $k \in \mathbb{Z}$, $\mu_{k}$ is the ideal luminance intensity in the k-th static scene and $\mathbf{S}_k(x, y, t)$ is the spike stream captured from k-th static scene.

\begin{figure}[t]
\includegraphics[width=\linewidth]{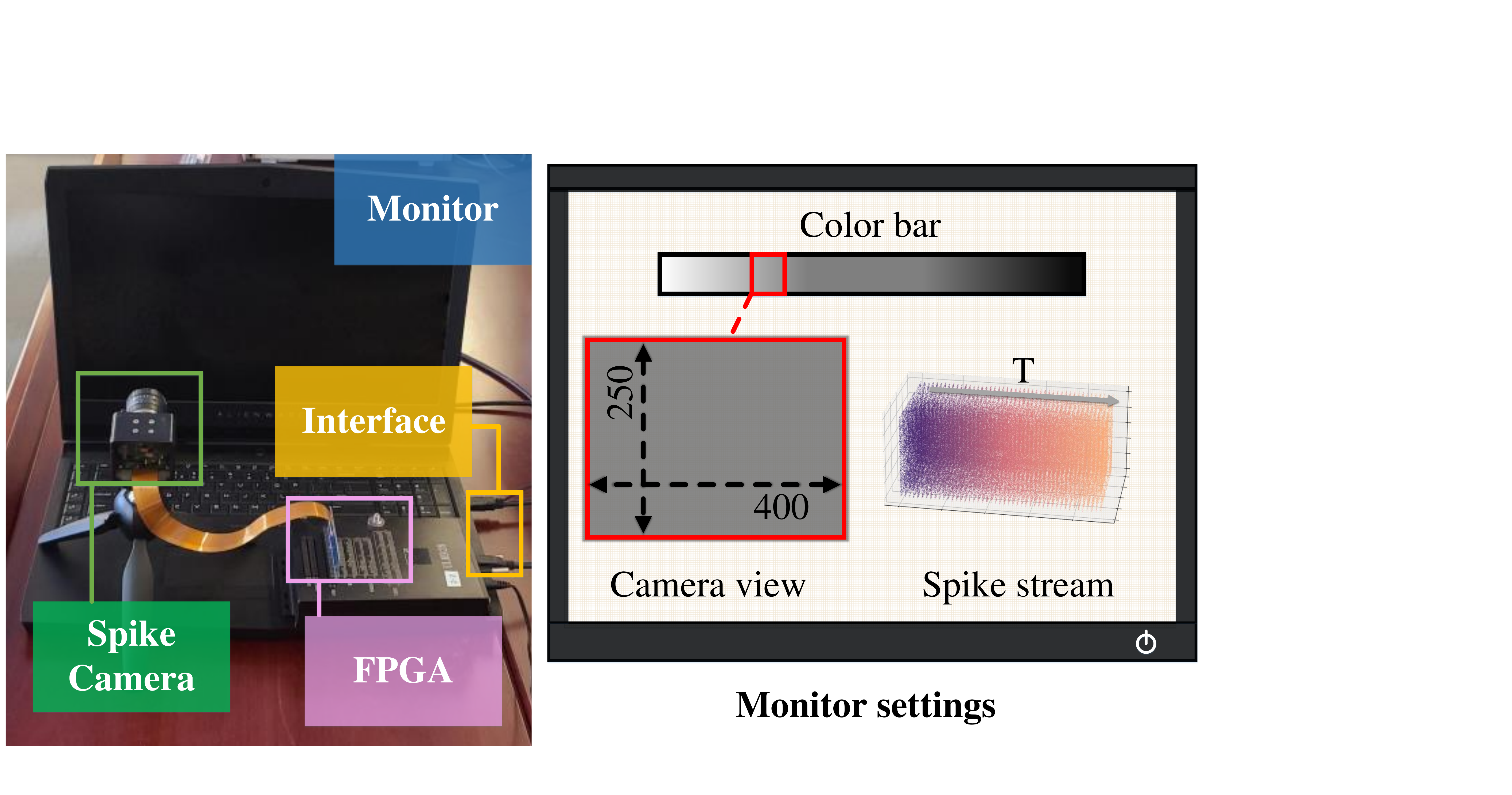}
\centering
\caption{Spike camera noise measurement setup.}\label{set}
\end{figure}
\begin{figure}[ht]
\includegraphics[width=\linewidth]{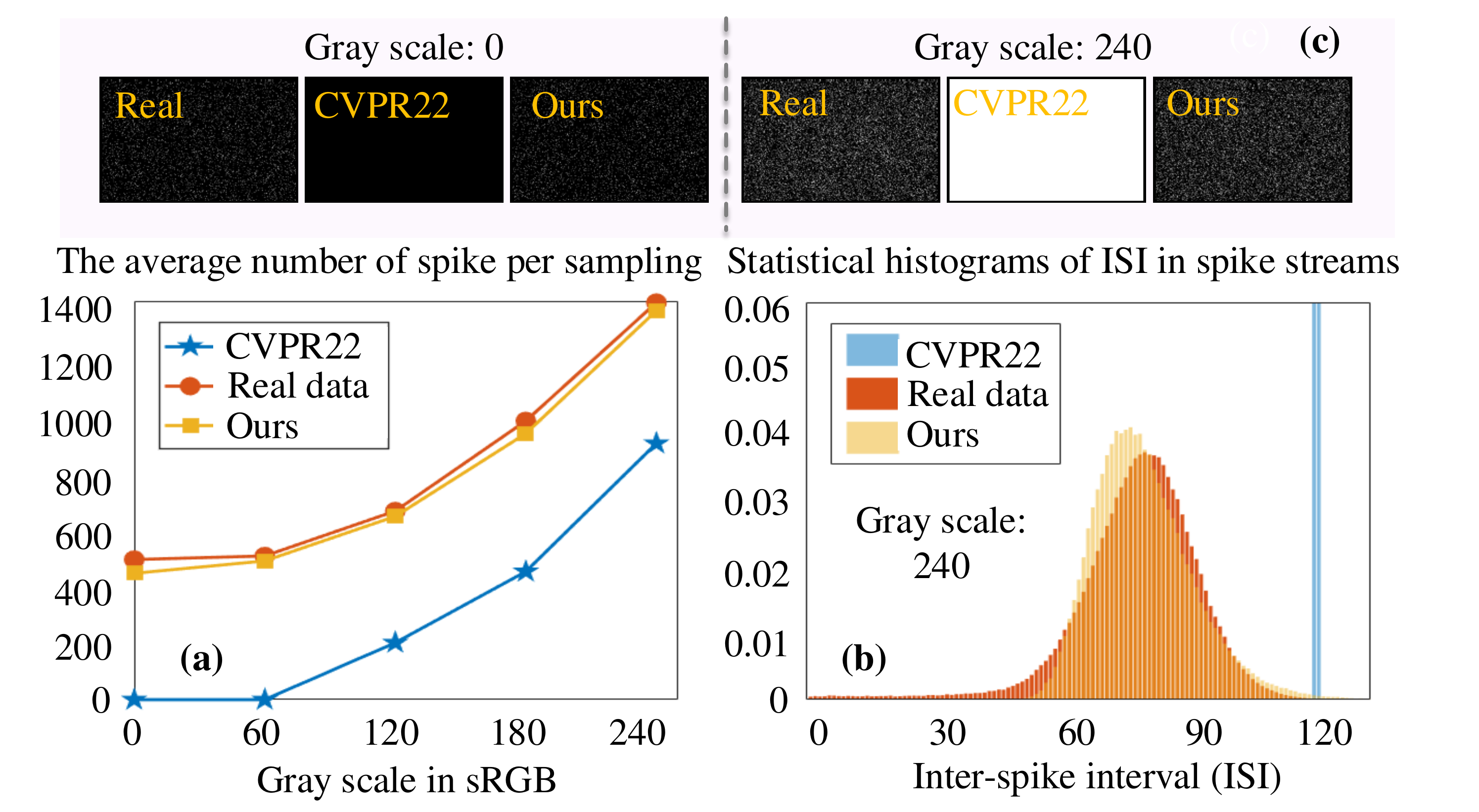}
\centering
\caption{Comparison of spike stream from \cite{2022scflow}, real spike camera and our model.(a) is the average number of spike per sampling.
(b) is histograms of Inter-spike interval (ISI) \cite{spikecamera} in spike stream. (c) is spike pattern where spike stream of $250 \times 400 \times 10$ is pressed onto a surface.
}\label{sim_ev}
\end{figure}
\hupar{Spike camera noise measurement setup}
As shown in Fig.~\ref{set}, we built the experimental scences where the spike camera shot computer monitor under 25 different grayscale backgrounds respectively. Further, accroding to (17), we can get a system of equations about $\alpha(x, y)$, $I_{\rm {dark}}(x, y)$, $C^S(x,y)$ and $V^S(x, y)$ for each pixel $(x, y)$. We can solve the above equations by optimization methods. Finally, the statics of spatial noise can be evaluated from $\alpha(x, y)$, $I_{\rm {dark}}(x, y)$, $C^S(x,y)$ and $V^S(x, y)$.

\hupar{Evaluation of simulated spike stream}
We compare simulated spike stream and captured real spike stream. Sampling scenes are solid backgrounds at 5 gray levels which are beneficial to ensure consistency between real and synthetic scenes. Sampling time is 0.1s and three kinds of spike stream are obtained under each scene. Finally, we have $5 \times 3$ spike stream. Fig.~\ref{sim_ev} illustrates reliability of SCSim.

\section{Method}
{Different from image and event denoising, spike stream denoising is to infer clean spike stream in a special spatio-temporal sampling from noisy spike stream.} Based on its unique data modality, we propose a tailored denoising framework.
In DnSS, denoised spike stream can be obtained by decoding inferred ISI sequence from noisy spike stream. Next, we use $\mathbf{S}^{t_1, t_2} \in {\{0, 1\}}^{H \cdot W \cdot (t_2 - t_1)}$ to denote a spike stream from time $t_1$ to $t_2$.
\subsection{Inter-spike Interval Plane}
Spike stream is not suitable to be directly used as supervisory signal because the encoded information is sparse (0 or 1).
Instead of spike stream, we first use its inter-spike interval plane, ISI \cite{spikecamera}, as supervision signal where the information of spike stream in a temporal window across time $t$ is compressed to a plane at time $t$ (written as $\mathbf{ISI}_t$).
\subsection{Network Architecture}
\begin{figure*}[thbp]
\includegraphics[width=\linewidth]{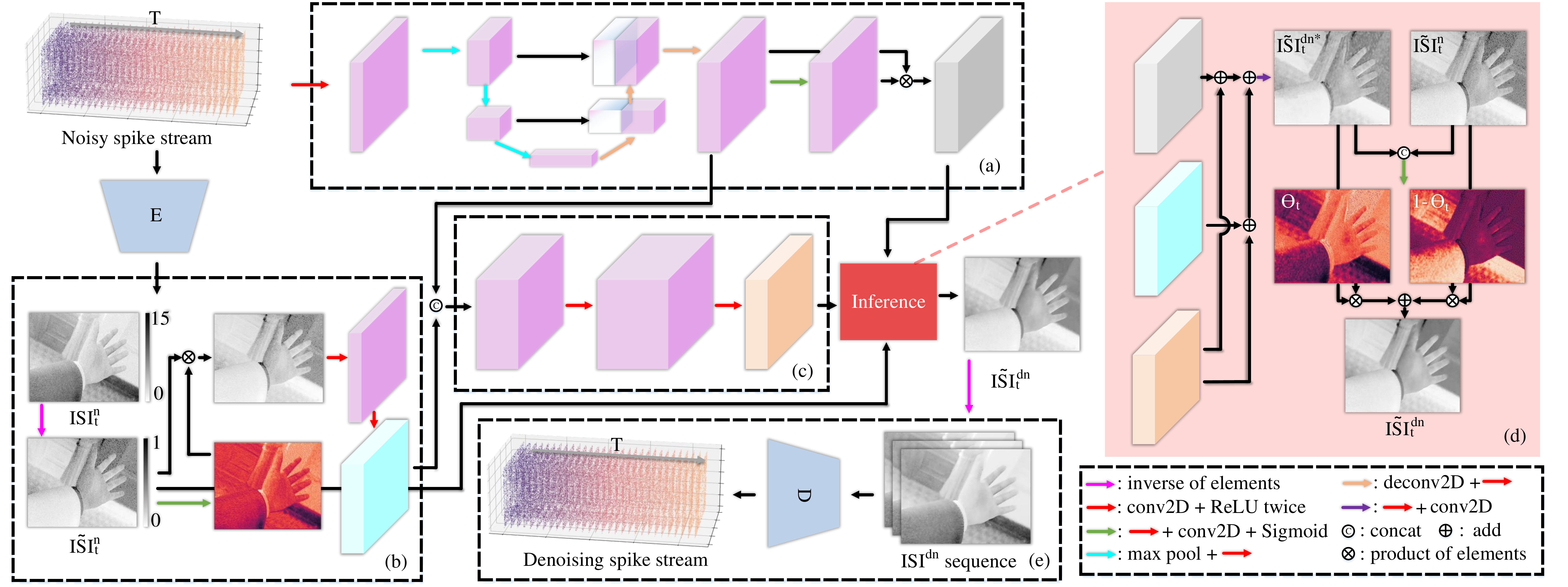}
\centering
\caption{The architecture of proposed denoising method for spike camera (DnSS). (a) Extract spatio-temporal denoising features of spike stream. (b) Extract texture features in spike stream. (c) Fuse spatio-temporal denoising features and texture features. (d) Infer ISI of clean spike stream based above features. (e) Decode inferred ISI sequence to denoised spike stream.}\label{DnSS}
\end{figure*}
\begin{figure}[thbp]
\includegraphics[width=\linewidth]{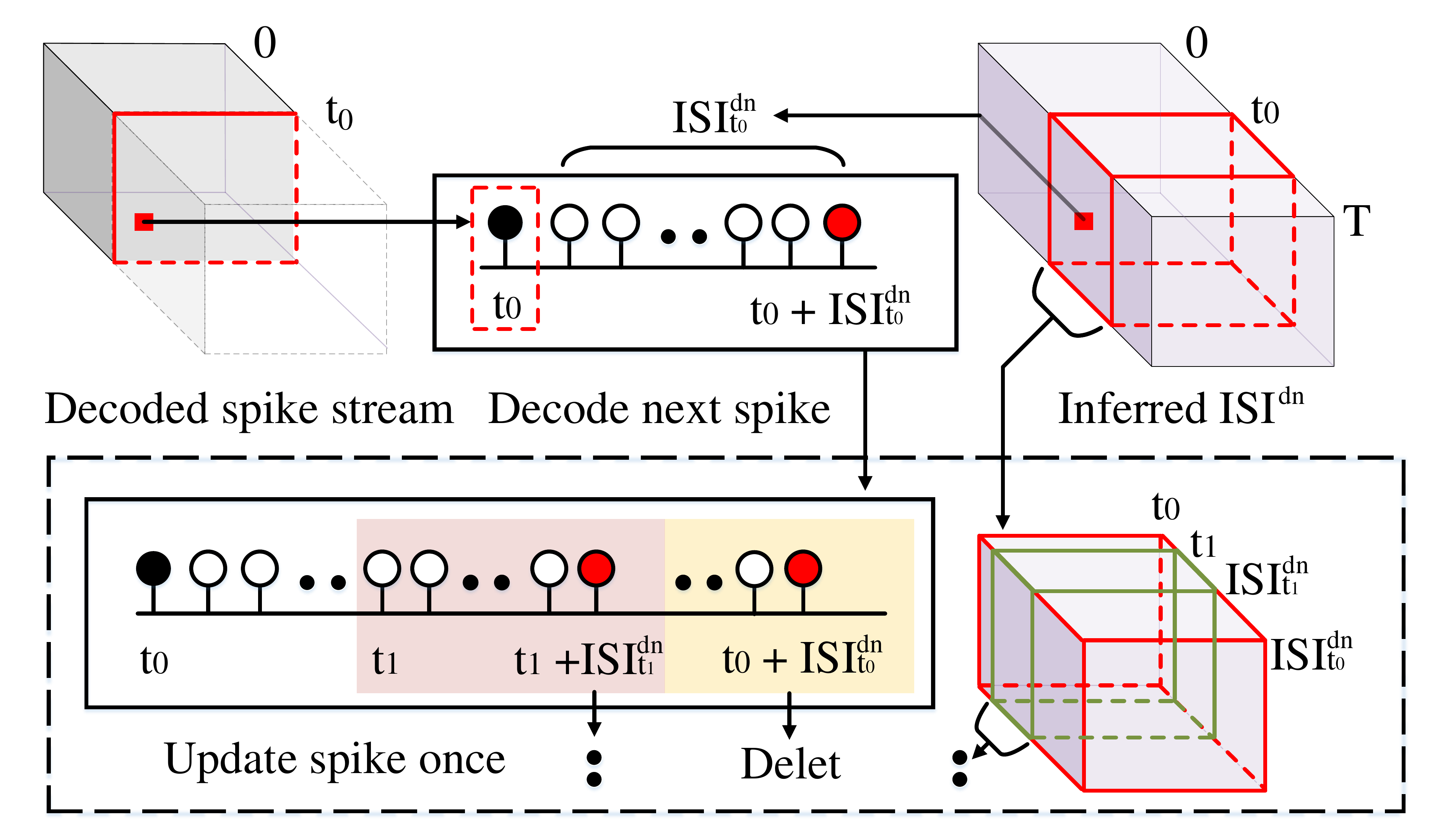}
\centering
\caption{The principle of multi-stage update strategy.}\label{decoder}
\end{figure}

As shown in Fig.~\ref{DnSS}, in DnSS, We use the noisy spike stream in a temporal window across time $t$, $\mathbf{S}_{n}^{t-\Delta t, t+\Delta t}$, to infer the normalized spike interval of clean spike stream at time $t$,  $\tilde{\mathbf{ISI}}_t^c$ where elements are mapped to $0-1$ by taking the inverse for each element in ${\mathbf{ISI}_t^c}$. 
Specifically, as Fig.~\ref{DnSS}(a), the input spike stream  $\mathbf{S}_{n}^{t-\Delta t, t+\Delta t}$ first is processed by 2-layer convolution where temporal noise is removed and spike information is compressed. Then, we can get the denoising feature map $\mathbf{F}_{dn}$ by using a U-Net and spatial attention module to remove spatial noise. 

In fact, all denoising methods are likely to introduce artifacts and
loss of details. Therefore, we design a refine module to restore high-frequency details in spike stream where $\tilde{\mathbf{ISI}}_t^n$ is as input. As Fig.~\ref{DnSS}(b), an spatial attention module is used to guide $\tilde{\mathbf{ISI}}_t^n$ to pay more attention to regions with rich textures. Then, we extract the texture feature map $\mathbf{F}_{tx}$ through 4-layer convolution. By introducing a fusion module, feature $\mathbf{F}_{tx}$ and feature $\mathbf{F}_{dn}$ can be fused as the fusion feature map $\mathbf{F}_{fu}$ as Fig.~\ref{DnSS}(c). Finally, as Fig.~\ref{DnSS}(d), we use previous feature maps to infer $\tilde{\mathbf{ISI}}_t^c$, i.e.,
\begin{align}
&\tilde{\mathbf{ISI}}_t^{dn*} = f(\mathbf{F}_{dn},\mathbf{F}_{tx}, \mathbf{F}_{fu}),\\
&[{\Theta}_t, \mathbf{1}-{\Theta}_t] = a([\tilde{\mathbf{ISI}}_t^{dn*}, \tilde{\mathbf{ISI}}_t^n]),\\
&\tilde{\mathbf{ISI}}_t^{dn} = {\Theta}_t \otimes \tilde{\mathbf{ISI}}_t^{dn*} + (\mathbf{1} - \Theta_t) \otimes \tilde{\mathbf{ISI}}_t^n,
\end{align}

where $f(\cdot)$ contributes the initial inference operation, $\tilde{\mathbf{ISI}}_t^{dn*}$ is the normalized spike interval of denoising spike stream at time $t$ which is first inferred, $a(\cdot)$ contributes is an attention operation, $\otimes$ is product of elements and $\tilde{\mathbf{ISI}}_t^{dn}$ is the normalized spike interval of denoising spike stream at time $t$. The $\tilde{\mathbf{ISI}}_t^{dn}$ is supervised by the normalized spike interval of clean spike stream at time $t$:
\begin{equation}
    \mathcal{L} = \Vert \tilde{\mathbf{ISI}}_t^{dn} - \tilde{\mathbf{ISI}}_t^{c} \Vert_2^2,
\end{equation}
where $\mathcal{L}$ is our loss function and $\Vert \cdot \Vert_2$ denotes the 2-norm.
After taking the inverse for each element in $\tilde{\mathbf{ISI}}_t^{dn}$, we can obtain denoised spike interval ${\mathbf{ISI}}_t^{dn}$. Due to the limitation of temporal window, the information in the input spike stream under dark light conditions may be insufficient and the denoised interval ${\mathbf{ISI}}_t^{dn}$ would be inaccurate. To this end, as shown in Fig.~\ref{decoder}, we design a multi-stage update strategy where, for pixel $(x, y)$, $\{{\mathbf{ISI}}_t^{dn}\}, t = t_0, t_0 + 1, \dots, t_0 + {\mathbf{ISI}}_{t_0}^{dn}(x, y)$ is used to update the final spike interval at time $t_0$, $\overline{\mathbf{ISI}}_{t_0}^{dn}$, i.e., 
\begin{equation}
\begin{aligned}
    &\overline{\mathbf{ISI}}_{t_0}^{dn}(x, y) = t_k + {\mathbf{ISI}}_{t_k}^{dn}(x, y) - t_0,
\end{aligned}
\end{equation}
where $k = \rm {argmax}_{m \in \mathbb{N^*}}\{\forall r \in \mathbb{N}, r \leq m, s.t., t_{r-1} < t_r < t_r +  {\mathbf{ISI}}_{t_r}^{dn}(x, y) < t_{r-1} + {\mathbf{ISI}}_{t_{r-1}}^{dn}(x, y)\}$. 
\begin{figure}[t!]
\includegraphics[width=\linewidth]{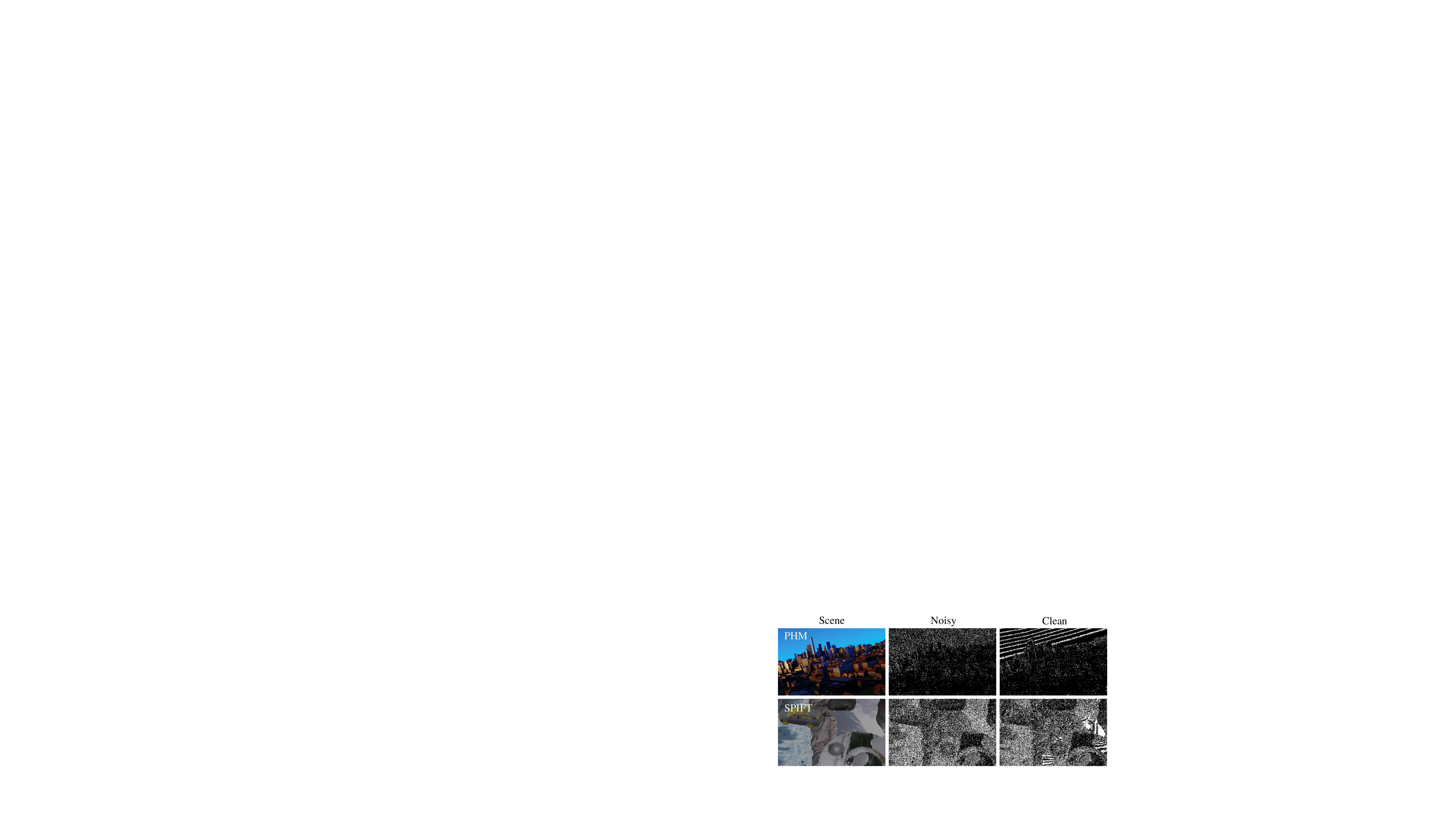}
\centering
\caption{Examples of noisy (clean) spike stream.}\label{data}
\end{figure}
Further, the final spike interval sequence can be decoded as denoised spike stream $\mathbf{S}_{dn}^{0, T}$.
\section{Experiments}
\subsection{Implementation Details}

\hupar{Datasets}
We propose noisy versions of SPIFT and PHM \cite{2022scflow}, $\text{SPIFT}_N$ and $\text{PHM}_N$. As shown in Fig.~\ref{data}, our datasets include simulated spike stream pairs by SCSim, i.e., clean (noisy) spike stream. We use $\text{SPIFT}_N$ including 100 random high-speed scenes as our training set while $\text{PHM}_N$ including 10 well-designed high-speed scenes as the test set. Besides, to evaluate the generalization of DnSS in real spike stream, we use the real spike stream \cite{rec3, rec1} captured from 5 high-speed scenes i.e., ``Car", ``Doll", ``Fan", ``Rotation1", ``Train".


\begin{figure*}[th!]
\includegraphics[width=\linewidth]{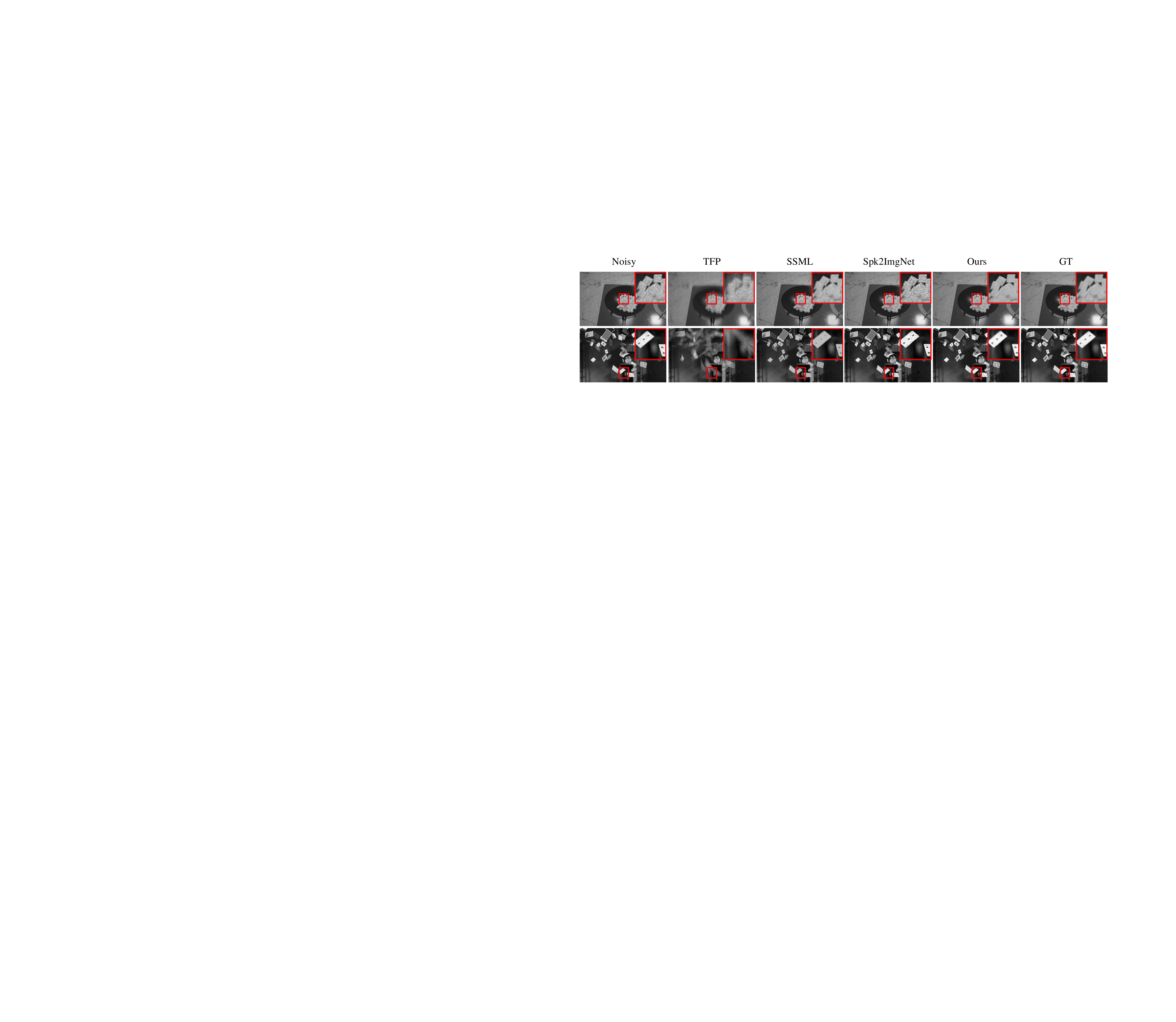}
\centering
\caption{Imaging results of denoised (noisy) spike stream on proposed dataset, $\text{PHM}_N$.}\label{r1}
\end{figure*}
\begin{figure*}[thbp]
\includegraphics[width=\linewidth]{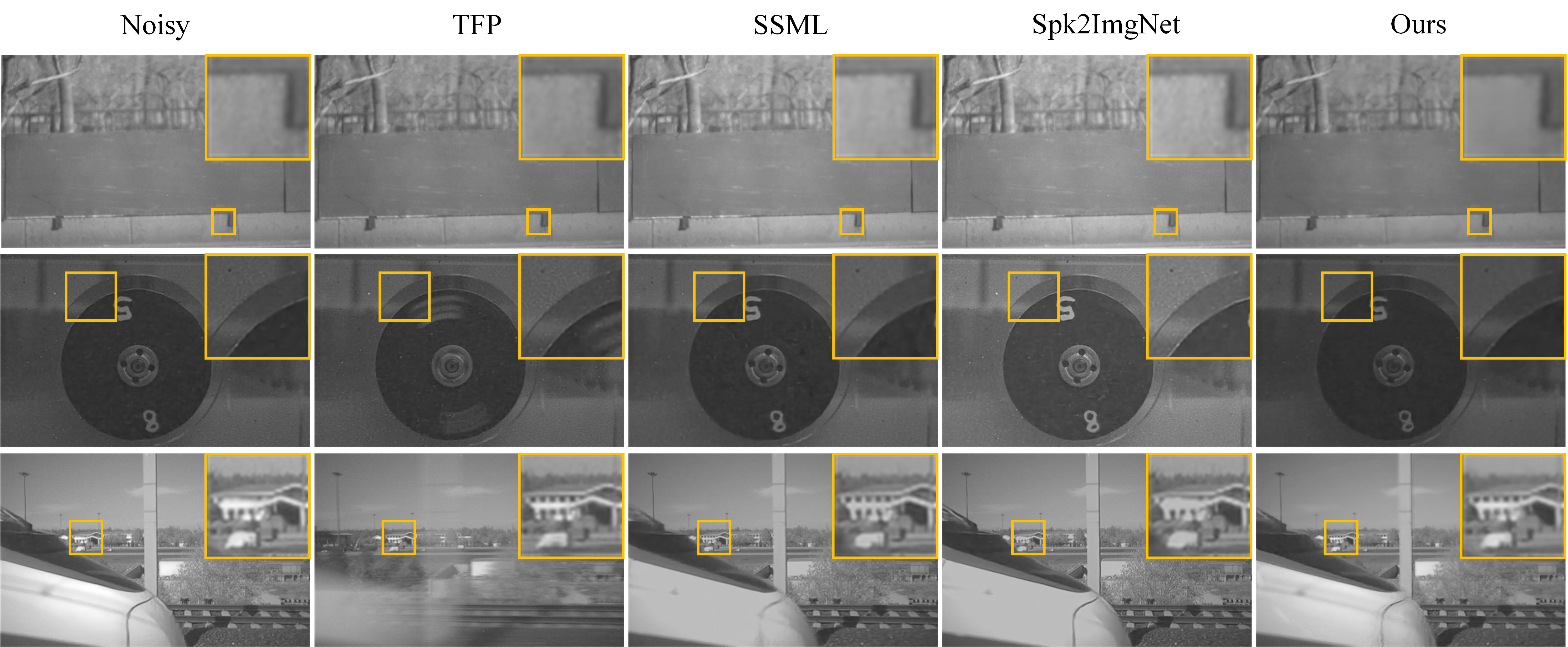}
\centering
\caption{Imaging results of denoised (noisy) spike stream on dataset from \cite{rec3,rec1} .}\label{r2}
\end{figure*}

\hupar{Train setup}
We implemented our DnSS in PyTorch. In training, Adam optimizer is adopted to optimize the networks and the initial learning
rate is set to 0.0001. DnSS is trained for 50 epochs with a batch size of 32 on 1 NVIDIA A100-SXM4-80GB GPU.

\hupar{Metric}
Since there are no methods to measure the similarity of spike stream, we cannot directly evaluate the performance of denoising methods. To this end, we compare imaging quality of different spike stream, i.e., PSNR and SSIM, where the state-of-the-art reconstruction method \cite{rec3} is used.

\subsection{Results}
We compare our method with the spike stream denoising pipeline where denoised image sequences are first obtained by reconstructing noisy spike stream, and then the denoised spike stream is synthesized from denoised image sequences with noise-free spike camera model. The performance of spike stream denoising pipeline depends on reconstruction methods. Here, we use the traditional method, TFP\cite{spikecamera}, the supervised method, Spk2ImgNet\cite{rec3} and the self-supervised method, SSML\cite{rec5} respectively.
\begin{table}[htbp]
  \centering
  \caption{Comparison of imaging results on $\text{PHM}_N$ dataset.}
    \begin{tabular}{lcccc}
    \toprule
            \tabincell{c}{Method} & \tabincell{c}{Param.} & \tabincell{c}{PSNR} & \tabincell{c}{SSIM} \\
    \midrule
    Noisy data  & /     & 27.70 & 0.698 \\
    TFP  & /     & 19.90  & 0.365 \\
    SSML & 2.38M & 27.08 & 0.723 \\
    Spk2ImgNet & 3.91M & 25.91 & 0.664 \\
    Ours  & \textbf{0.27M} & \textbf{29.61} & \textbf{0.847} \\
    \bottomrule
    \end{tabular}%
  \label{tab_r1}%
\end{table}

\hupar{Qualitative Evaluation on Synthetic Datasets.}
we compare the imaging performance of noisy spike stream and denoised spike stream. As shown in table.~\ref{tab_r1}, the denoised spike stream from TFP and Spk2ImgNet cannot improve image reconstruction quality and SSML can remove noise in spike stream to a certain extent. Our denoising method gets the best average performance.
Moreover, the parameters of DnSS are far less than SSML and Spk2ImgNet.
Fig.~\ref{r1} shows the imaging results of denoised spike stream by different methods. TFP can introduce a lot of motion blur during denoising. Although texture details can be retained well by Spk2ImgNet, noise cannot be effectively eliminated. Comparing the first two, SSML is able to retain partial textural details during denoising. However, the denoising performance of SSML is limited. Our method can effectively remove the noise in noisy spike stream without the loss of texture details.

\begin{table}[htbp]
  \centering
  \caption{Comparison of imaging results on dataset from \cite{rec1,rec3}}
    \begin{tabular}{lrrrrr}
    \toprule
    Metric & \multicolumn{1}{l}{Noise} &
    \multicolumn{1}{l}{TFP} & \multicolumn{1}{l}{SSML} & \multicolumn{1}{l}{Spk2ImgNet} & \multicolumn{1}{l}{Ours} \\
    \midrule
    NIQE($\downarrow$) & \multicolumn{1}{c}{4.07} & \multicolumn{1}{c}{4.29} & \multicolumn{1}{c}{4.10} & \multicolumn{1}{c}{4.32} & \multicolumn{1}{c}{\textbf{3.76}} \\
    \bottomrule
    \end{tabular}%
  \label{tab_r2}%
\end{table}%

\hupar{Qualitative Evaluation on Real Datasets.}
We compare the imaging quality of denoised spike stream by different methods are as input. Due to absence of GT in spike camera datasets recording high-speed scenes, we use a no-reference IQA metric, i.e., NIQE to evaluate different denoising methods in Table.~\ref{tab_r2}. Besides, the imaging results of different denoising methods are shown in Fig.~\ref{r2}.  We can find our method can effectively remove noise introduced by hot pixels which proves that our denoising method has pretty generalization on real data. 

\subsection{Ablation}
To investigate the effect of the proposed refine module (RF) and multi-stage update strategy (MUS), we compare three baseline methods. Baseline1 is the basic baseline without RF or MUS. Based on Baseline1, Baseline2 uses multi-stage update strategy to reduce the inference error and Baseline3 adds refine module to improve texture details.
\begin{table}[htbp]
  \centering
  \caption{Evaluation of the proposed modules on $\text{PHM}_N$ dataset.}
    \begin{tabular}{lccccc}
    \toprule
    Method & \multicolumn{1}{l}{Param.} & \multicolumn{1}{l}{MUS} & \multicolumn{1}{l}{RF} & \multicolumn{1}{l}{PSNR} & \multicolumn{1}{l}{SSIM} \\
    \midrule
     Baseline1     & \textbf{0.20M} & \ding{53}     & \ding{53}     & 28.67 & 0.822 \\
     Baseline2     & \textbf{0.20M} & \ding{51}     & \ding{53}     & 29.13 & 0.828 \\
     Baseline3     & 0.27M & \ding{53}     & \ding{51}     & 29.11 & 0.841 \\
     Ours     & 0.27M & \ding{51}     & \ding{51}     & \textbf{29.61} & \textbf{0.847} \\
     \bottomrule
    \end{tabular}%
  \label{tab_ab1}%
\end{table}%
\begin{figure}[thbp]
\includegraphics[width=\linewidth]{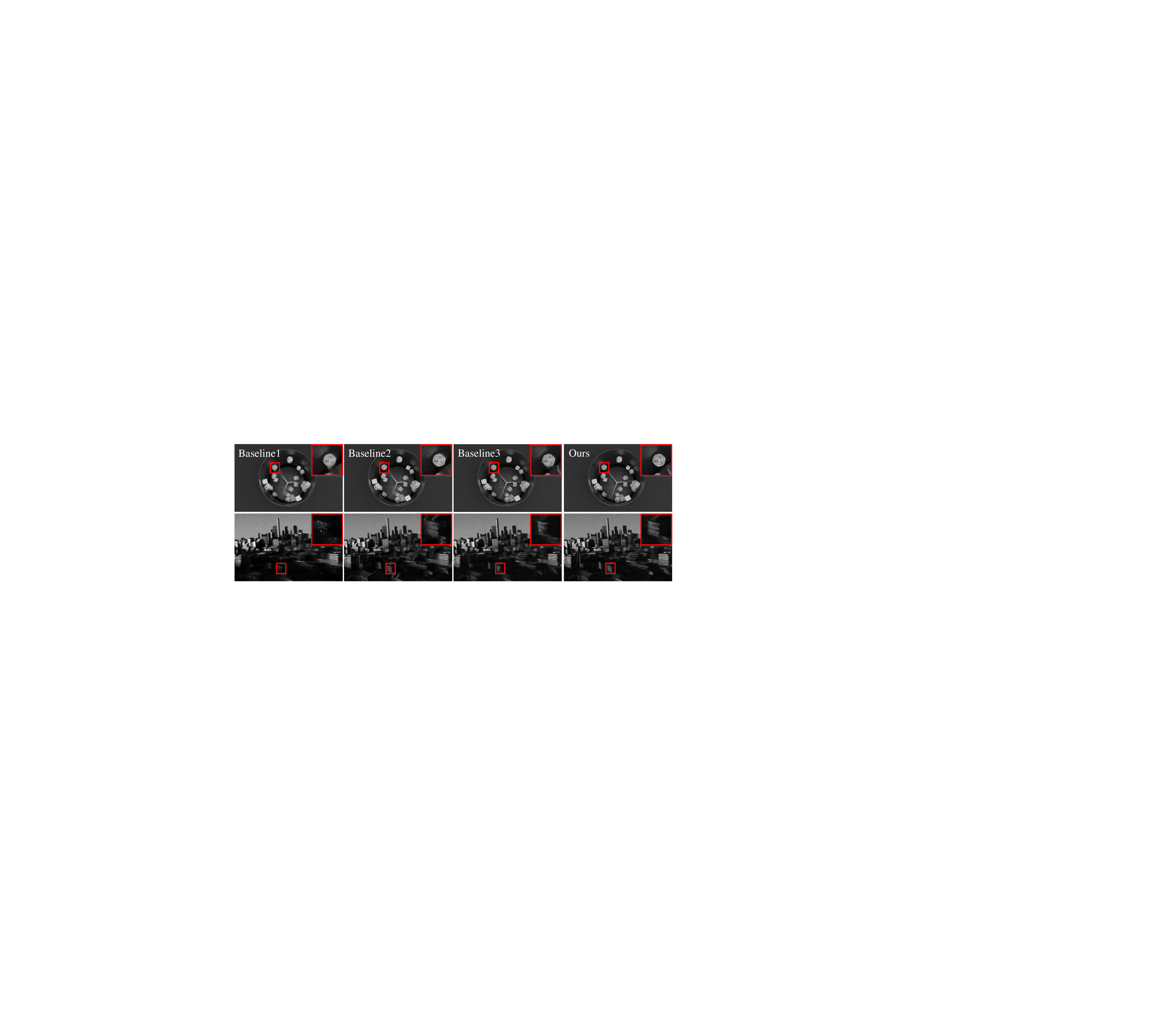}
\centering
\caption{Ablation study of the proposed modules on $\text{PHM}_N$.}\label{ab1}
\end{figure}

\begin{table}[htbp]
  \centering
  \caption{Evaluation of the proposed modules on dataset from \cite{rec1,rec3}.}
    \begin{tabular}{lrrrr}
    \toprule
    Metric & \multicolumn{1}{l}{Baseline1} &
    \multicolumn{1}{l}{Baseline2} & \multicolumn{1}{l}{Baseline3} & \multicolumn{1}{l}{Ours} \\
    \midrule
    NIQE($\downarrow$) & \multicolumn{1}{c}{4.15} & \multicolumn{1}{c}{4.14} & \multicolumn{1}{c}{3.80} &  \multicolumn{1}{c}{\textbf{3.76}} \\
    \bottomrule
    \end{tabular}%
  \label{tab_ab2}%
\end{table}%
\begin{figure}[thbp]
\includegraphics[width=\linewidth]{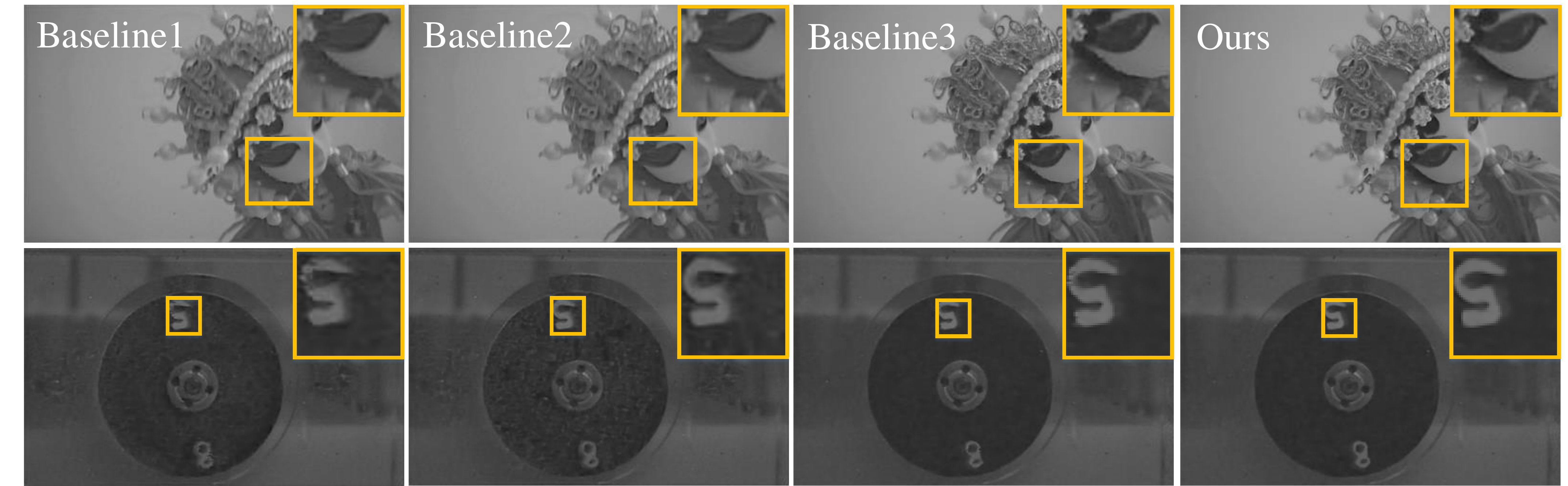}
\centering
\caption{Ablation study of the proposed modules on dataset from \cite{rec1,rec3}.}\label{ab2}
\end{figure}

\hupar{Evaluation on Synthetic Datasets.}
As shown in Table.~\ref{tab_ab1}, we find that RF and MUS can improve the performance of the basic baseline. By combining RF and MUS,  Our DnSS outperforms the basic baseline by around 1 dB. Fig.~\ref{ab1} shows the imaging results of denoised spike stream.

\hupar{Evaluation on Real Datasets.}
As shown in Table.~\ref{tab_ab2}, Our DnSS achieves the best denoising performance on the real spike stream by introducing RF and MUS. The imaging results in Fig.~\ref{ab2} also show advantages of RF and MUS intuitively. Specifically, by using the MUS module into Baseline1, the wrong interval in Baseline2 can be updated which makes the bad points in dark eliminated. By using the RF module into Baseline1, the edge details become smoother. Combining RF and MUS, Our DnSS can keep the details in the real scene better.
\section{Conclusion}
We propose a tailored and lightweight denoising framework for spike camera, DnSS, where the ISI in clean spike stream can be inferred based on RF, and then we obtain the denoised spike stream by using MUS to decode the inferred ISI. To evaluate the performance of denoising methods, a spike camera simulation method, SCSim, is proposed where we model and measure the noise in spike camera refer to its unique circuit. Based on SCSim, we synthesize the spike-based denoising datasets, $\text{SPIFT}_N$ and $\text{PHM}_N$. The experiment demonstrates DnSS can denoise spike stream efficiently. Moreover, DnSS can be generalized well on real spike stream.
{\small
\bibliographystyle{ieee_fullname}
\normalem
\bibliography{scflow}
}
\end{document}